\documentclass[11pt]{article}

\usepackage[T1]{fontenc}
\usepackage{lmodern}
\usepackage{microtype}
\usepackage[margin=1in]{geometry}
\usepackage{amsmath,amssymb,mathtools}
\usepackage{graphicx}
\usepackage{xcolor}
\usepackage{framed}
\definecolor{shadecolor}{gray}{0.95}
\usepackage{hyperref}
\hypersetup{
  colorlinks=true,
  linkcolor=blue!40!black,
  citecolor=blue!40!black,
  urlcolor=blue!40!black
}

\usepackage[backend=biber, style=numeric-comp, sorting=none]{biblatex}
\title{Likelihood scoring for continuations of mathematical text: a self-supervised benchmark with tests for shortcut vulnerabilities}
\author{Daniel Ranard\\
Department of Physics, California Institute of Technology, Pasadena, CA 91125, USA}
\date{\today}

\newcommand{\cliptwo}{\ensuremath{\mathrm{clipLL}_{2}}}
\newcommand{\boxheading}[1]{\par\smallskip\noindent\textbf{#1.}\par\nobreak\vspace{0.3em}}
\newcommand{\Zpred}{Z_{\mathrm{pred}}}
\newcommand{\CB}{C_B}
\newenvironment{bracket-display}{\[}{\]}

\providecommand{\mathscr}[1]{\mathit{#1}}

\begin{document}

\maketitle

\begin{abstract}
We introduce an automatically generated benchmark for predicting hidden text in
technical papers.
A paper supplies visible context \(X\) and a hidden continuation \(Y\); the
evaluated model writes an auxiliary forecast string \(Z\), and a separate
scorer assigns next-token probability to \(Y\) both with and without conditioning on \(Z\).
This gives a label-free test of whether \(Z\) transmits information about the
continuation, compared against controls where \(Z\) is recent context rather than a forecast.

Our main testbed is equation-suffix prediction: the
predictor sees context and the first part of a displayed equation,
then forecasts the rest.
The task mixes surface-level arXiv/\TeX{} text modeling with reasoning-sensitive
inference; the suffix is one of many roughly equivalent continuations,
so the benchmark is read statistically rather than item-by-item.
On 1363 equation continuations from 138 recent physics and mathematics papers,
forecasts from GPT-5.5, Opus 4.7, and GPT-5.4 nano all improve (clipped) likelihood
over the context control under both Qwen3-8B and Kimi K2.6
scorers, distinguishing model families and reasoning-effort settings without
human labels.

To emulate shortcuts where \(Z\) further primes the scorer rather than
making a useful forecast, we also fine-tune the scorer on context-only
prompts and apply it to held-out papers as a stronger control. GPT-5.5
forecasts still beat this fine-tuned control; GPT-5.4 nano forecasts do not.
Longer prose/\TeX{} continuations show positive but noisier lift over
controls, concentrated near the beginning of the target.
These results support cross-model likelihood scoring as a static benchmark and
as a setup for probing shortcut vulnerabilities before reinforcement learning
or model-selection optimization is applied.
\end{abstract}

\section{Introduction}

Technical papers offer a natural source of prediction tasks: guessing the next
lines of a derivation, equation, or proof. Tasks of this form are abundant and
reference-grounded, which makes them an interesting target for evaluating
language models. But any one continuation is hard to judge cleanly, since the
true text is just one realization among many roughly equivalent forms.

This paper asks whether automatically generated tasks of this kind can serve as
a useful benchmark. A paper supplies visible context \(X\) and a hidden
continuation \(Y\). The model under evaluation writes an auxiliary string \(Z\),
intended to help predict \(Y\). A separate language model, used only as a
likelihood scorer, then assigns next-token likelihood to the true \(Y\) under
prompts with and without \(Z\). The benchmark asks whether \(Z\) improves this
likelihood, and compares the forecast against controls that replace \(Z\) with
nearby context of equal length rather than a genuine prediction.

We also report tests probing shortcut vulnerabilities: whether the score can be
improved by context-stuffing or scorer-priming rather than by useful forecasts.
These tests may inform whether such a task could eventually serve as a
reinforcement-learning reward, although we do not perform such training here.

A useful way to situate this kind of task, at least informally, is between two
poles. At one pole is raw next-token prediction, the pre-training objective. It
is abundant and automatic---any text corpus produces it for free---but most
tokens do not particularly reward extended reasoning. For the typical token,
predictive gains come from modeling increasingly fine-grained surface details
of the text distribution: which synonym an author prefers, notational
conventions, and so on. One could deploy extended reasoning on such tokens, but
the resulting gains may not be especially interesting, whether considered as a
benchmark of reasoning ability or as a post-training reward. At the other pole
are hand-crafted, verifiable technical problems---competition mathematics, code
with unit tests, textbook exercises---of the kind that drive much of
recent training via reinforcement learning with verifiable rewards (RLVR)~\cite{lambert2024tulu3,shao2024deepseekmath,deepseek2025r1}.
While these tasks demand reasoning, they are relatively scarce and narrow. It
seems especially hard to obtain abundant, verifiable problems that are embedded
deeply within a domain: a calculation that arises in a current physics or
mathematics manuscript often requires more diffuse background and more
reasoning under uncertainty than a synthetic exam item.

Our main testbed is what we call equation-suffix prediction: given a paper's
preceding context and some prefix of a displayed equation, the model predicts
the rest of the equation. This sits somewhere between the two poles. The task
inherits some of the abundance and automaticity of next-token prediction---any
equation-rich technical paper supplies many examples---while also more
plausibly benefiting from longer reasoning. The size of the prediction target
also lands in a useful range. Very short targets are noisy: a model that reasons
correctly but produces one of several valid next-token predictions
can be marked wrong on essentially a coin flip, wasting the signal from a
correct insight. Very long targets are noisy too, but for different
reasons: the space of valid continuations grows quickly, and the intended
scoring becomes ill-determined. An equation suffix of moderate length (tens to hundreds of characters here) may mitigate both failure modes. Drawing tasks from recent manuscripts is also
motivated by contamination concerns familiar from fresh-benchmark work~\cite{livebench2024,antileakbench2025,arxivroll2025}.

This paper has two related purposes. The first is to present a static
benchmark: an automatically generated, recent-arXiv equation-suffix dataset on
which different predictor models and reasoning-effort settings can be compared
via likelihood lift under fixed scorers, without human labels. Our second purpose is to take some preliminary steps toward asking whether the resulting signal could in
principle be used as a reward for RLVR-style
post-training~\cite{deepseek2025r1} in technical domains. We do not run
reinforcement learning here. Instead, we audit the static signal against context controls, including ones with fine-tuned scorers, which probe one specific class of non-forecast strategies that a predictor undergoing RL might exploit. The audit is partial, and we flag what it does and
does not cover. Concerns of this kind are increasingly familiar from the
reward-hacking and verifier-gaming
literature~\cite{gao2023overoptimization,gamingVerifiers2025}.

We emphasize that the score is derived from the probabilities of an
autoregressive language model, not an LLM-as-a-judge rubric evaluator producing
free-form judgments. Related likelihood-improvement signals have recently been
proposed as rewards for long-form generation~\cite{gurung2025longform},
next-word prediction~\cite{shen2025bow}, next-turn dialogue
prediction~\cite{gandhi2026dialogue}, and reasoning training based on
future-token prediction~\cite{zelikman2024quietstar, dong2025rpt, li2025rlpt, hatamizadeh2025rlp},
and for evaluating LLMs without using an LLM as a judge~\cite{xu2025gem};
we discuss positioning relative to these works in Section~\ref{sec:related}.

This likelihood-scoring route is a middle ground between two more obvious
evaluation choices. Asking an instruction-tuned model to judge which forecast
is better would be flexible, but it would also introduce the familiar
LLM-as-judge vulnerabilities: rubric drift, style preference, and direct
promptability of the
evaluator~\cite{zheng2023judging,liu2024narcissistic,panickssery2024recognize,zhao2025onetoken}. Simple string or
embedding similarity would be less flexible, but risks being too brittle for
equations and for technical continuations with many approximately equivalent forms.

\paragraph{Contributions.}
We define an automatically generated benchmark for technical continuations, in which
predictor-written auxiliary strings are evaluated by likelihood lift under fixed
autoregressive scorers (Section~\ref{sec:likelihood}). We instantiate this idea
with a recent-arXiv equation-suffix dataset (Section~\ref{sec:equation-benchmark})
and show that it separates model families and reasoning-effort settings under
both Qwen3-8B~\cite{qwen3report} and Kimi K2.6~\cite{moonshot_k26_2026,kimi2025k2} scorers
(Section~\ref{sec:equation-results}). We
study the scoring mechanism itself, including catastrophic-token behavior and
softened scoring rules (Section~\ref{sec:likelihood} and
Appendix~\ref{app:metric-sensitivity}). We audit the likelihood signal against
context-only controls of increasing strength, finding that GPT-5.5 forecasts
survive the strongest control while GPT-5.4 nano forecasts do not
(Section~\ref{sec:controls}). We also contrast this cleaner equation-suffix regime
with longer mixed prose/\TeX{} continuations, where lift remains positive but
becomes noisier and more scaffold-sensitive
(Section~\ref{sec:longer-continuations}).

\section{Likelihood lift given an auxiliary string}
\label{sec:likelihood}

We now formalize the central quantity of our benchmark, the per-token
likelihood lift, and the scoring rule built from it.

Our setup involves two language models: the model under evaluation,
sometimes referred to as the predictor, and a secondary model used
only for likelihood scoring. We write \(J\) for the scorer, an autoregressive language model, and
\(p_J(Y \mid P)\) for the conditional probability \(J\) assigns to a
continuation \(Y\) after a prompt \(P\); this factorizes as
\(p_J(Y \mid P) = \prod_{t=1}^T p_J(y_t \mid P, y_{<t})\) over the
tokens of \(Y = (y_1, \ldots, y_T)\), with \(y_{<t} = (y_1, \ldots, y_{t-1})\). Given visible context \(X\), auxiliary string \(Z\) (typically written
by the predictor), and hidden target continuation \(Y\), our basic quantity of interest is the
per-token increase in scorer log-likelihood that comes from including
\(Z\),
\[
  \Delta_J(X,Z,Y)
  =
  \frac{1}{T}\log p_J(Y\mid X,Z)
  -
  \frac{1}{T}\log p_J(Y\mid X).
\]
We refer to \(\Delta_J\) as the (per-token, log-)\emph{likelihood
lift} of \(Z\): the increase in average scorer log-likelihood that
the auxiliary string buys for the true continuation. (Roughly, we are
trying to maximize the mutual information between forecast \(Z\) and
continuation \(Y\), conditional on the partial context \(X\).) More generally, we also use ``likelihood lift'' for analogues where the
second term \(\log p_J(Y\mid X)\), the scorer's likelihood without
\(Z\), is replaced by its likelihood under some other control condition. The notation suppresses a
deterministic prompt scaffold that presents \(X\), then \(Z\), then a
marker returning the scorer to the paper just before \(Y\); the
scaffold is fixed across all conditions in a comparison, so changing
the auxiliary string is the only difference between
\(\log p_J(Y\mid X,Z)\) and \(\log p_J(Y\mid X)\). Note that \(X\)
refers to what appears in the scorer's prompt. The predictor that
produces \(Z\) typically sees a broader context (e.g., 10,000 characters
of preceding paper text in Section~\ref{sec:equation-benchmark}); this
broader context shapes \(Z\) but is not given directly to the scorer.
For the equation-suffix benchmark, the scorer prompt has the schematic form
\begin{quote}
\small
\begin{shaded}
\begin{verbatim}
% First equation:
\begin{<env>}
<equation prefix><Z>
\end{<env>}

% Same equation:
\begin{<env>}
<equation prefix>
\end{verbatim}
\end{shaded}
\end{quote}
\noindent and the scored target appended after this prompt is the true suffix \(Y\),
followed by the closing display delimiter. We tried small variants of this
scaffold in preliminary exploration, with qualitatively similar results.

In practice we compare a predictor-generated forecast string
\(\Zpred\) to matched controls, rather than only to scoring without \(Z\). The most important control is the same-budget recent-context
control \(\CB\) (where the subscript \(B\) denotes the character budget):
instead of receiving a model forecast, the scorer
receives raw source text immediately preceding \(X\), with the same
character budget available to \(\Zpred\). This directly tests a
natural degenerate strategy: use the auxiliary string for nearby
context rather than for a prediction of the hidden continuation.

Raw mean log-likelihood is informative but can be dominated by rare severe
local mismatches, especially in \TeX{}. We therefore use a softened
headline metric, \cliptwo{}. With
\(\lambda_t = \log p_J(y_t \mid X, Z, y_{<t})\) the per-token log-likelihood, we define
\[
  \cliptwo \;:=\; \frac{1}{T}\sum_{t=1}^{T} \max(\lambda_t, -2).
\]
That is, each per-token log-likelihood is clipped from below at $-2$, then
averaged. Larger \cliptwo{} indicates better forecasts. This still rewards making target tokens likely, while
preventing a small number of catastrophic tokens from erasing an
otherwise useful forecast.

The choice of \cliptwo{} is motivated both empirically and mechanistically. In preliminary
token-level inspections, we noticed that raw log-likelihood was often
dominated by very large penalties on a small number of tokens. As one
recurring pattern, when a forecast matched the true equation closely
for several tokens, the scorer's predictive distribution over the
next token sharpened, so a subsequent deviation --- sometimes itself
a mathematically equivalent or otherwise valid continuation ---
incurred a much larger log-loss than the same deviation would in
isolation, and could dominate the per-token average even when the
forecast remained useful on later tokens. Across small exploratory
samples and related tasks, several softened variants of the log-loss mitigated this behavior in a similar way.
We use \cliptwo{} as a fairly simple representative: it clips many catastrophic token
contributions while leaving most token contributions unchanged. We use
\emph{softness} to refer to how aggressively a scoring rule bounds per-token
losses; \cliptwo{} is a relatively soft score, while raw log-likelihood is the
unsoftened limit. Throughout the paper, ``lift'' or ``likelihood lift''
(without qualification) refers to the \cliptwo{} version of \(\Delta_J\)
unless otherwise specified; alternative softenings are named explicitly when used. Appendix~\ref{app:metric-sensitivity} shows that the
main qualitative findings of this paper hold across a range of bounded
scoring rules. As the score is made less soft, the weakest GPT-5.4 nano
settings begin to show the same catastrophic-token behavior seen under raw LL,
illustrating the mechanism above.
Appendix~\ref{app:equation-order-probe} illustrates this
behavior on toy examples of mathematically equivalent continuations.

\section{What the equation-suffix prediction task measures}

As mentioned in the introduction, equation-suffix prediction mixes at least two kinds of skill. First, the task has
a technical language-modeling component: a useful predictor must track
local arXiv/\TeX{} surface conventions and the kind of expression an
author is likely to write next. Second, it has a more directly reasoning-sensitive
component: if the visible context determines a bound, recurrence,
variational identity, or next derivation line, a model that can infer
or derive that structure should often write a better forecast. 

The likelihood-lift score is an imperfect metric ---
for a single hidden suffix, two forecasts may both be partially right
in different ways, with no canonical verdict on which is closer to the
paper. But it is also not arbitrary: a frozen scorer's likelihood is more
reproducible and less rubric-driven than a free-form judge's decision
about which forecast looks better. On many examples, especially when
comparing a strong predictor's forecast to a much weaker one's,
one continuation is visibly much closer than the
other; for subtler comparisons, such as adjacent reasoning-effort
settings of the same model, item-level ordering is intrinsically noisy.

The benchmark's claim is therefore statistical: averaged over many
automatically generated examples, forecasts from stronger predictors and
higher reasoning settings make the actual held-out suffixes more
likely. This empirical ordering is evidence that the benchmark
contains reasoning-sensitive signal, even though that signal is mixed
with technical-text modeling. Extra reasoning may help by explicitly
deriving or constraining the hidden equation, or by improving surface-level
modeling of technical \TeX{}. Either mechanism is interesting for
technical-continuation modeling; this benchmark by itself does not
decompose those mechanisms, and doing so would require further
targeted experiments.

\section{Equation-suffix dataset}
\label{sec:equation-benchmark}

The main benchmark consists of 1363 equation-suffix \emph{cuts} from 138 recent
arXiv manuscripts in the \texttt{quant-ph}, \texttt{hep-th}, and
\texttt{math-ph} categories (reflecting the author's familiarity rather than a
systematic field selection). A cut is produced by slicing inside a displayed
equation; the remaining suffix is the prediction target. Sources were selected
using criteria applied before any scoring: recent papers, technical and
equation-rich style,\footnote{In practice, this combined paper length
(\(\geq 25\) pages) with automated keyword filters favoring research papers
and (somewhat arbitrarily) matching the author's interests, fixed before any
scoring.} \TeX{} source availability, and successful source/cut validation. The construction is deliberately automatic. We did not
hand-pick favorable equations or reject papers after inspecting model outputs;
the scripts admit papers by simple source criteria, generate candidate cuts, and
then score the resulting fixed benchmark cuts.

For each cut, the predictor is prompted with the
previous 10,000 characters of paper context and the visible prefix of
the displayed equation, then asked to continue the equation from exactly where
it stops, in roughly the target length or less, writing only the continuation.
Predictor generations used sampling temperature 1.0.\footnote{This is the documented default for both the OpenAI Responses API and the Anthropic Messages API, and we did not override it.}
The cut location is chosen by a simple syntactic rule: scan displayed equations,
require at least 10,000 preceding context characters, choose a relation or
operator site in the middle third of the equation body, and use the remaining
suffix as \(Y\), with \(50 \leq |Y| \leq 400\) characters before the closing
display delimiter. This rule intentionally does not select for derivations or interestingness; it is a simple way to produce many technical prediction tasks.

The dataset is also constructed to avoid trivial redundancy. At most 10 cuts
are selected from any paper, and at most one cut is taken from any displayed
equation; and the 1363 cuts have 1363 unique displayed-equation keys.\footnote{Cuts where $Y$ appears verbatim in the predictor's preceding context are excluded as a pre-scoring filter (17 cuts in total); partial substring matches are common and retained, since reusing earlier expressions is part of how technical papers are written.}
Thus repeated-paper examples are distinct equation occurrences, not overlapping
sliding windows through the same suffix.

The scorer \(J\) is one of two open-weight language models with publicly
accessible token-level log-probabilities: Qwen3-8B~\cite{qwen3report} from
Alibaba (8.2B parameters, dense) or Kimi K2.6~\cite{moonshot_k26_2026,kimi2025k2}
from Moonshot AI (1T-parameter Mixture-of-Experts, 32B activated per token). We picked these two because they sit at very different scales --- a
small dense model and a large mixture-of-experts --- and come from
unrelated vendors. If both scorers tell the same story, the signal is
less likely to be an artifact of one family~\cite{liu2024narcissistic}. Both were accessed via the
Fireworks AI \texttt{/completions} endpoint in raw prompt-completion
logprob mode.

Appendix \ref{app:expanded-prompt-example} shows two expanded examples, including
the predictor and scoring prompts. Appendix \ref{app:equation-examples} shows
five randomly sampled examples, with the true paper suffix and the GPT-5.5
(high reasoning) forecast displayed side by side.

\section{Main equation-suffix results}
\label{sec:equation-results}

\begin{figure}[!htbp]
\centering
\includegraphics[width=\linewidth]{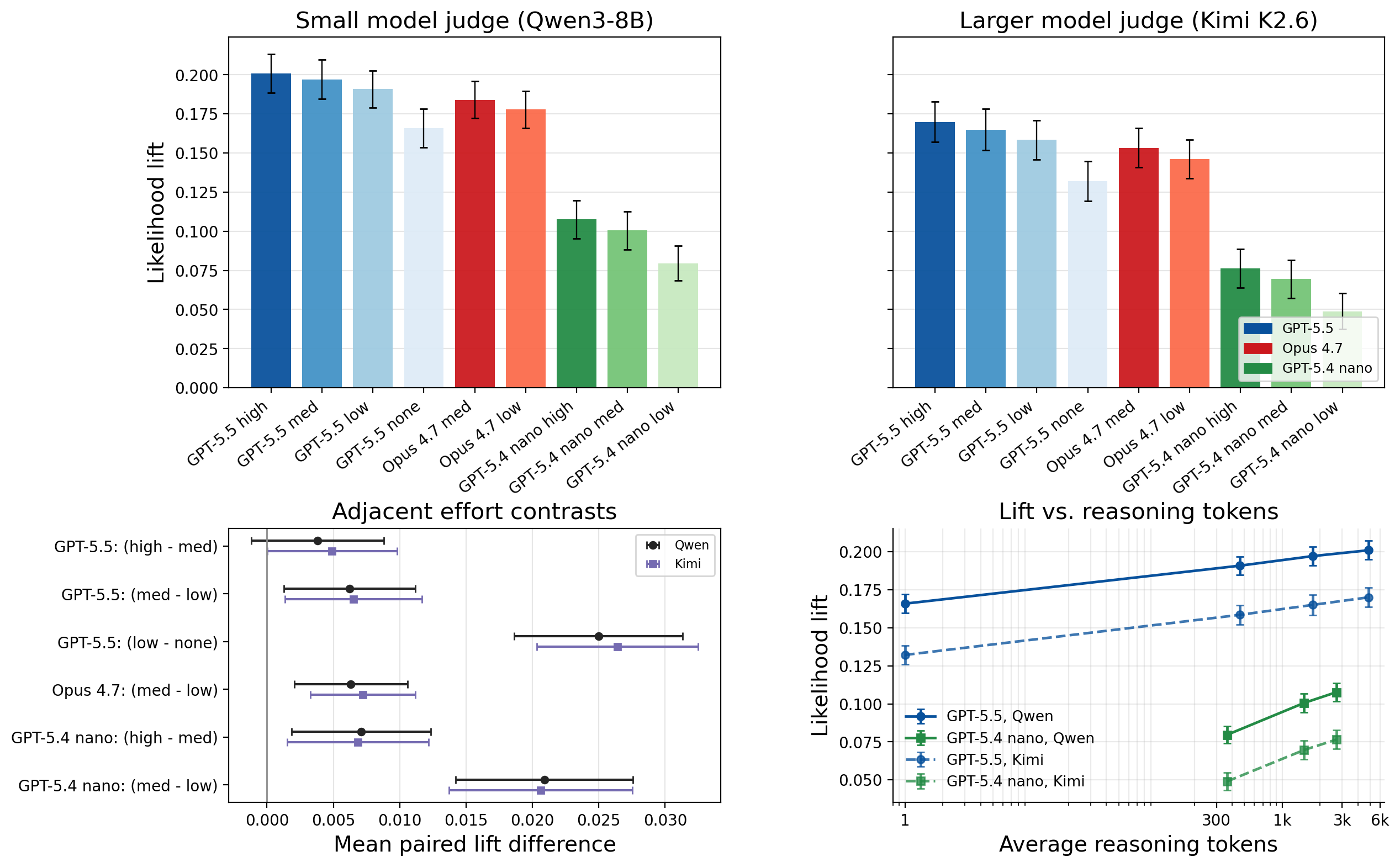}
\caption{Equation-suffix benchmark. Top: forecast lift over the same-budget
recent-context control for Qwen3-8B and Kimi K2.6 likelihood scorers. Bottom
left: paired adjacent reasoning-effort contrasts on the same equation cuts.
Bottom right: forecast lift versus average API-reported hidden reasoning-token
use, for the OpenAI predictor settings only (see Table~\ref{tab:generation-usage-diagnostics}). Likelihood lift is reported using \cliptwo{} per
target token. The bottom-right panel uses a logarithmic \(x\)-axis, with
zero-reasoning settings plotted at \(x=1\). Error bars show approximate 95\%
intervals (\(\pm 2\) paper-clustered standard errors). All adjacent contrasts shown are
significant at \(p<0.05\) under a normal approximation using the
paper-clustered SE, except GPT-5.5 (high reasoning) minus GPT-5.5 (medium
reasoning) under the Qwen scorer.
Numerical values are in Tables \ref{tab:fig1-lift-values} and
\ref{tab:fig1-adjacent-values}.}
\label{fig:equation-benchmark-main}
\end{figure}

All predictor settings beat the same-budget context control \(\CB\) on the combined
1363-cut benchmark under both fixed scorers. The effect is largest for OpenAI's GPT-5.5~\cite{openai_gpt55_2026}
predictor settings, intermediate for Anthropic's Claude Opus 4.7~\cite{anthropic_opus47_2026},
and smaller for OpenAI's GPT-5.4 nano~\cite{openai_gpt54nano_2026}.
The benchmark also detects reasoning-effort gradients, where reasoning effort
is the API generation setting used by the predictor when writing \(Z\).
Within each predictor family, we see a monotone reasoning-effort ordering:
GPT-5.5 high $>$ medium $>$ low $>$ none; Opus 4.7 medium $>$ low; GPT-5.4
nano high $>$ medium $>$ low. This ordering holds under all five bounded
scoring rules we tested (Appendix~\ref{app:metric-sensitivity}), on both scorers. While individual
adjacent contrasts have varying significance (Figure~\ref{fig:equation-benchmark-main}, bottom left), the joint robustness of the
monotone pattern across metrics and scorers is itself illustrative.
Figure~\ref{fig:equation-benchmark-main} (bottom right) plots lift against
API-reported reasoning-token use for the OpenAI predictor settings.
While finer reasoning-effort sweeps would be needed to confirm a precise
functional form, the diminishing returns visible here are at least
consistent with the logarithmic dependence of performance on test-time
compute often seen for reasoning models more broadly
\cite{snell2024scaling}. The GPT-5.5 (no reasoning) point,
plotted at $x=1$, also appears consistent with this trend.
While these effort levels are provider-defined API settings rather than
calibrated units of computation, the API does report hidden reasoning-token
counts, which we use as the compute axis in
Figure~\ref{fig:equation-benchmark-main} (bottom right). Representative
usage statistics are in Table~\ref{tab:generation-usage-diagnostics}.

\begin{figure}[!htbp]
\centering
\includegraphics[width=\linewidth]{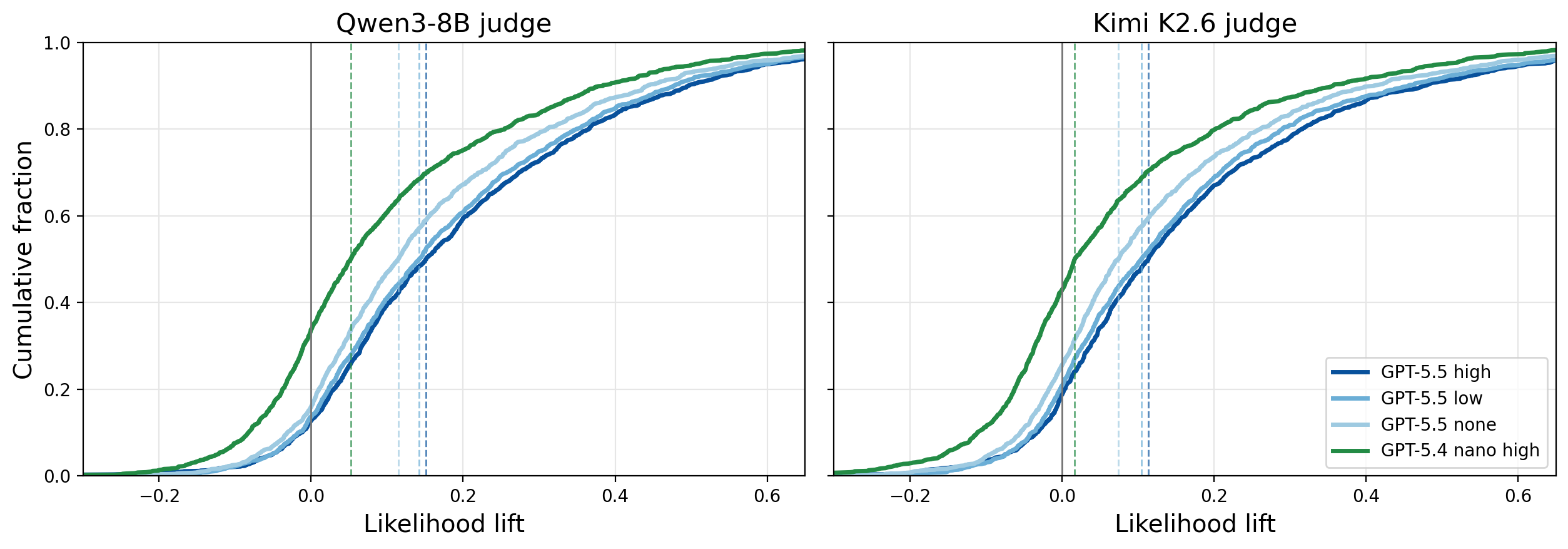}
\caption{Per-cut forecast-lift distributions for selected predictor settings.
Each curve is the empirical cumulative distribution of forecast lift
over the same-budget recent-context control; dashed vertical lines mark medians.
GPT-5.4 nano (high reasoning) versus the GPT-5.5 settings shows a clear
distributional shift. The GPT-5.5 (high reasoning) versus GPT-5.5 (low
reasoning) contrast is smaller, but it is still visible
through the median region rather than only as a few extreme right-tail wins.}
\label{fig:equation-lift-distribution}
\end{figure}

The main interpretation is that forecast strings contain information about
hidden technical continuations beyond what is recovered by spending the same
budget on recent source context. The replication across two different scoring
models reduces the chance that the signal is only a quirk of one model family.
Figure \ref{fig:equation-lift-distribution} shows that the larger separations
are not merely tail effects: GPT-5.5 settings are broadly right-shifted relative
to GPT-5.4 nano (high reasoning), and GPT-5.5 (high reasoning) and GPT-5.5
(low reasoning) are broadly right-shifted relative to GPT-5.5 (no reasoning).

The finest adjacent effort comparisons are weaker. In particular, GPT-5.5
(high reasoning) has a higher mean lift than GPT-5.5 (medium reasoning) under
both scorers, and the model-level lift medians are also ordered in this
direction. But this is still
the least robust adjacent contrast: it is the only adjacent comparison in Figure
\ref{fig:equation-benchmark-main} that is not significant at the
two-standard-error level under one scorer, and under Qwen the paired
high-minus-medium median is zero while GPT-5.5 (high reasoning) beats GPT-5.5
(medium reasoning) on fewer than half of cuts. This clarifies the benchmark's
intended use: it orders models and settings by average lift over many
automatically generated cuts, not by adjudicating each individual continuation.
The adjacent high-reasoning--medium-reasoning result should therefore be
read as the smallest step in a robust monotone gradient, not as a clean
item-wise ordering.

\section{Controls for context-only shortcuts}
\label{sec:controls}

This section describes stronger probes of the likelihood lift produced by the forecast string, beyond the same-budget control. The same-budget context control is an initial audit for reward-hackability, but it
does not exhaust possible shortcuts. (We use ``shortcut'' in the broad sense of learning shortcuts~\cite{geirhos2020shortcut}.) We therefore also study stronger static (non-adversarially-optimized) controls. The most severe current control uses supervised fine-tuning (SFT): we train
a LoRA~\cite{hu2021lora} (Low-Rank Adaptation) adapter for the Qwen3-8B scorer on a context-only route (with no
forecast string), evaluated on papers disjoint from the fine-tuning papers. This control is not an intended reward scorer; it is a
stress test for whether forecasts can beat a learned context-only shortcut.

We use this control because a future predictor optimized against a likelihood
reward might discover non-forecast uses of the auxiliary string \(Z\). The
most obvious one is to spend the auxiliary budget on nearby source context. In
the equation benchmark, the same-budget control \(\CB\) gives the scorer the
previous \(B=|Y|+40\) source characters before the equation, followed by the
equation prefix, instead of a forecast.\footnote{Note that $C_B$ replaces the entire scaffold (the scorer reads source context and the equation prefix directly) rather than just substituting source context for $Z$ inside the scaffold. A pilot variant that instead placed the same source context as $Z$ inside the scaffold did not improve over the empty-scaffold baseline; we therefore use the unscaffolded $C_B$ as our main control.} The triple-budget variant gives three
times that much recent context.

It is useful to be precise about what the context-only SFT control tests. It is
aimed at the predictor taking non-forecast shortcuts: raw context stuffing (filling \(Z\) with raw paper context), reordered or
restructured nearby source text, and reusable format or scaffold priming that
helps steer the scorer into a better task-conditional state. The control is
deliberately strong for this purpose. It is not merely a longer context
baseline, but a many-parameter adaptation of the scorer trained directly on the
\cliptwo-based objective on papers disjoint from evaluation. A clever
\(Z\) may prime the scorer by perturbing its input; SFT primes the scorer by
perturbing its weights. SFT is generally the more flexible of the two
(gradient search rather than handcrafting), so this control should capture
a broad class of generic context-use and scaffold-priming gains that a
bounded \(Z\) string would otherwise have to emulate.

Important routes remain outside this stress test. It does not test
forecast-plus-hacking strategies, where a predictor keeps transmitting real
information about \(Y\) while also adding scorer-specific priming. Nor does
it test highly idiosyncratic adversarial strings with no clean weight-space
analogue --- that is, strings exploiting scorer behavior in ways no natural
fine-tuning of the scorer would capture. Thus
GPT-5.5 beating this control is evidence against a non-forecast context-or-scaffold shortcut explaining the signal, not a complete reward-hackability proof.
Testing the remaining attack surface would require optimizing \(Z\) directly
against a frozen scaffold-aware scorer.

\begin{figure}[!htbp]
\centering
\includegraphics[width=\linewidth]{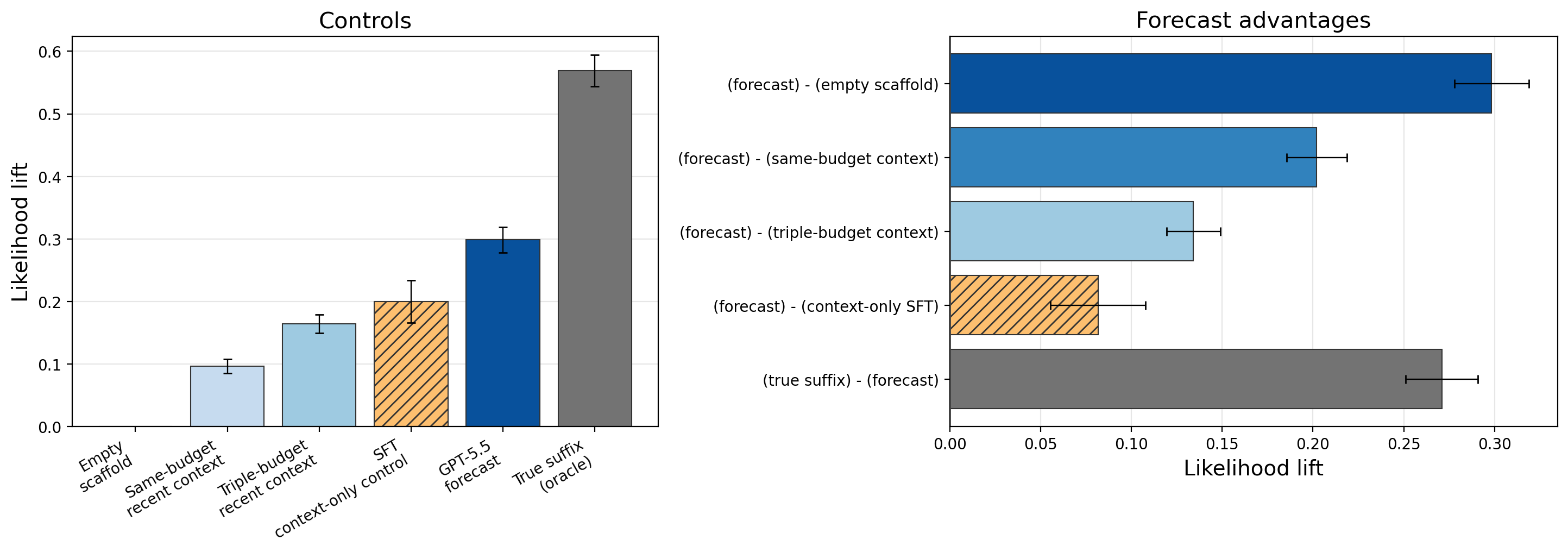}
\caption{Static controls for equation suffixes with GPT-5.5 (high reasoning) forecasts and
a Qwen3-8B likelihood scorer. The ladder compares an empty scaffold,
same-budget context,
triple-budget context, a context-only SFT control evaluated on held-out papers,
the model forecast, and a true-suffix upper bound (where $Z$ is the true $Y$ itself). The right panel reports
paired forecast advantages over these controls. Error bars show approximate 95\% intervals
(\(\pm 2\) paper-clustered standard errors). Numerical values are in Tables
\ref{tab:fig2-control-ladder-values} and
\ref{tab:fig2-forecast-control-values}.}
\label{fig:equation-static-controls}
\end{figure}

That said, the scoring scaffold constrains this remaining attack surface. The likelihoods extracted from the scorer are the same next-token probabilities that define its autoregressive continuations, so the scaffold can be interpreted by asking what continuation behavior it induces. The \emph{First equation / Same equation} scaffold is designed to make copying the completed first equation the natural induced behavior: after seeing one completed equation, the scorer is placed at the start of the same equation again and scored token by token on the true continuation. Appendix~\ref{app:equation-order-probe} illustrates this behavior in toy examples: exact forecasts achieve \cliptwo{} scores around \(-0.01\), close to the maximum possible \cliptwo{} value of \(0\). The true-suffix condition in Figure~\ref{fig:equation-static-controls} therefore provides an oracle reference point for the full benchmark: it measures how much lift is obtained when the auxiliary string literally contains the hidden suffix \(Y\) in the intended scaffold.

This does not make the true-suffix condition a formal supremum over all possible strings \(Z\). A sufficiently pathological scorer/scaffold interaction could in principle improve the presentation of the answer or exploit scorer-specific behavior. But a non-forecast shortcut that does not transmit example-specific information about \(Y\) should not systematically approach this oracle reference on a heterogeneous set of hidden suffixes. Indeed, strong forecasts are already near the top of the clipped scoring range on a meaningful fraction of cuts: for GPT-5.5 (high reasoning) under the Qwen3-8B scorer, 16.3\% of cuts have \cliptwo{} at least \(-0.10\). Thus a successful attack would have to compete not only with the mean forecast lift, but also with many examples where the forecast is already close to the scaffold's effective maximum under the clipped metric.

On held-out papers, GPT-5.5 forecasts still beat this context-only SFT control
under \cliptwo, whereas GPT-5.4 nano forecasts do not. This is a useful
separation: strong predictors transmit information that survives an unusually
hard static shortcut control, while weaker predictors do not.

\section{A harder regime: longer technical continuations}
\label{sec:longer-continuations}

We also test longer mixed prose/\TeX{} continuations. This setting is closer to
broad technical forecasting, but less clean: the continuation is longer, many
surface forms are plausible, and likelihood scoring depends more strongly on
the scoring model and the length of the scored target.

The prose/\TeX{} experiment is constructed separately from the equation-suffix
benchmark. It uses 661 cuts from 40 recent papers. A cut here splits a paper
into a visible context and a hidden continuation target, with the split
chosen between balanced \TeX{} blocks rather than inside an equation
environment. No scored cut occurs inside an explicit \TeX{} environment,
although the visible context and the target often contain displayed
equations. In this dataset, 405 of the 661
targets contain equation-like material after the cut.

By default, the predictor sees 10,000 characters of preceding context and is
asked for roughly 1,800 characters of continuation. For scoring, the
forecast \(Z\) is truncated to a 1,000-character budget and embedded in a
scaffold of the form
\begin{quote}
\small
\begin{shaded}
\begin{verbatim}
[last 2000 characters before Y]
% Notes about what's next:
% <Z>
% Returning to paper text:
[same 2000 characters before Y]
\end{verbatim}
\end{shaded}
\end{quote}
\noindent with the scored target \(Y\) appended after this prompt. The simple
recent-context control is unscaffolded: the scorer reads a contiguous
3,000-character pre-target context (the 2,000 characters immediately before
the cut, plus 1,000 extra preceding characters matching the \(Z\) budget),
then \(Y\). This is the same budgeting logic as in the equation suffixes:
a forecast should beat a control that spends its whole \(Z\)-budget on
recent context.

\begin{figure}[!htbp]
\centering
\includegraphics[width=\linewidth]{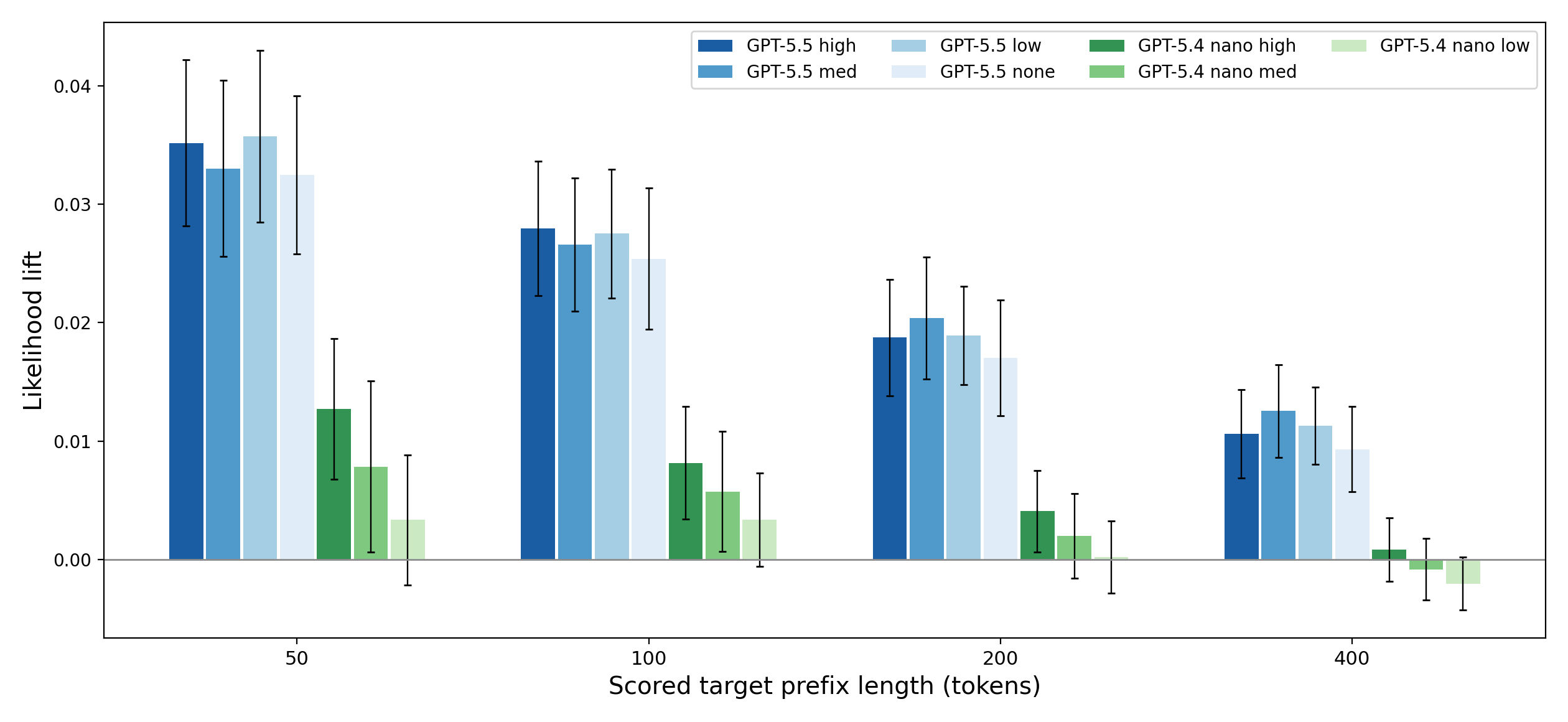}
\caption{Prose/\TeX{} continuation forecast lift by scored target window. The
same generated forecast strings are scored on the first \(N\) tokens of the
true continuation. Lift is strongest near the beginning of the target and decays
as the scored continuation length increases. Likelihood lift is reported using
\cliptwo{} per target token. The \(x\)-axis is categorical, with each group
corresponding to a scored prefix length \(N\); error bars show approximate 95\% intervals
(\(\pm 2\) paper-clustered standard errors). Numerical values are in Table
\ref{tab:fig3-prose-window-values}.}
\label{fig:prose-window-lift}
\end{figure}

The longer-continuation experiment shows positive forecast lift, especially over
the first part of the target. It also clearly separates GPT-5.5 from GPT-5.4 nano.
However, adjacent GPT-5.5 reasoning-effort ordering is weaker and noisier than
in equation suffixes. We interpret this as evidence that long technical
continuations are a real but harder regime, where target length, scorer design,
and scoring softness become central.

We also ran SFT comparisons in this longer-continuation setting. For
equation suffixes, we did not use a scaffold-aware SFT scorer in the headline
comparison because the unscaffolded scorers without SFT already gave a clean positive signal
against the strong controls. Leaving the scorer untuned keeps the result simpler and more
portable across scorer and predictor choices, since no scorer-specific
fine-tuning is required. For prose/\TeX{} continuations the signal is weaker,
and the forecast scaffold (without a fine-tuned scorer) is easier to beat with an adversarially
trained context route. In that setting, scaffold-aware SFT is a natural choice: the reward designer may tune the scorer on papers disjoint from the
evaluation papers so it understands the intended \(X,Z,Y\) interface, and then freeze it before
evaluation or any hypothetical predictor optimization.

Concretely, one Qwen3-8B LoRA is trained on the intended forecast scaffold using
GPT-5.4 nano forecast strings on training papers disjoint from evaluation;
another is trained on the direct 3,000-character recent-context route. Both use the same
loss corresponding to \(\cliptwo\), plus a small raw negative-log-likelihood
penalty (to give nonzero gradient on the clipped region). The forecast-scaffold SFT represents a possible scaffold-aware frozen
scorer: it removes confusion about the X-Z-Y interface before evaluation. The
recent-context SFT is an adversarial control: it asks how well a context-only
strategy could do after the same kind of clipped-objective training.
In this comparison, forecast-scaffold SFT beats the recent-context SFT under
\cliptwo{} most clearly for short scored prefixes; on raw (unclipped)
log-likelihood it does not, since the SFT objective itself is clipped, so
neither model was trained to do well on the catastrophic tokens that
dominate the raw score. More broadly, the prose/\TeX{} regime is noisier and more sensitive to methodological choices than equation suffixes; we treat this SFT comparison as exploratory rather than as a headline benchmark.

\section{Related work}
\label{sec:related}

Relevant prior work includes continuation-prediction benchmarks,
likelihood-based rewards, generative evaluation, and reward-hackability
analysis. To organize our comparison to other work, we recall a few key
features of the present setup.

\begin{enumerate}
\item The predictors are reasoning models that can do extensive internal
chain-of-thought before producing $Z$. The scorer sees only $X$ and $Z$,
never the predictor's internal reasoning $W$. We assess the likelihood
lift induced by $Z$; the literal log-likelihood is computed on $Y$,
but our quantity of interest is the increase in $\log p_J(Y \mid X, Z)$
over a baseline. Note that if this signal were used as a reinforcement-learning
reward, $W$ would feel optimization pressure only indirectly: the chain
adapts insofar as different reasoning produces a more useful $Z$. This
is gentler than a setup that reads $W$ and rewards it directly, which
would steer the chain's content and form toward what a particular
scorer prefers.

\item The scorer is a fixed model distinct from any predictor, so
likelihood lifts from different predictors are measured against the same
baseline. This is what makes the benchmark a common scale on which
different predictor models can be compared.

\item While we do not train models here, we run a few explicit controls,
aimed at various ways the signal could be gamed. Using a separate, frozen
scoring model makes the signal reproducible and decoupled from the RL
dynamics under potential training.

\item Likelihood lift is computed from a per-token clipped/softened
log-likelihood (\cliptwo{}), not raw sum-log-prob; in our setting raw
scoring is often dominated by a small number of catastrophic per-token
losses that softened variants control.

\item The domain is technical paper continuations from recent arXiv
manuscripts, perhaps a desirable target for reasoning.
\end{enumerate}

\paragraph{Likelihood-based benchmarks.}
A small group of LLM benchmarks use likelihood as the evaluation
signal instead of an LLM judge; our setup is in that group. (Older
continuation-prediction benchmarks such as LAMBADA and HellaSwag~\cite{paperno2016lambada,zellers2019hellaswag}
establish the evaluation format but do not use an auxiliary string $Z$
or a likelihood-improvement signal.) GEM~\cite{xu2025gem}
(building on earlier work in information
elicitation~\cite{kong2018eliciting,lu2024eliciting}) estimates pointwise
mutual information between two reports about a shared object, relying on
a conditional-independence assumption between the reports. Sharma et
al.~\cite{sharma2025rawcorpora} extract domain-specific target terms
(keywords and TF-IDF-ranked vocabulary) from corpora and score by the
predictor's probability of the target term. We differ in scoring a predictor-written forecast $Z$ by
its lift in scorer log-likelihood of the actual hidden continuation
$Y$ --- multi-token, clipped per-token, budget-matched against a
control. GEM also discusses shortcut filters for superficial
correlations; our audit instead targets context stuffing, format
priming, and scaffold exploitation via budget-matched and
fine-tuned-scorer baselines.

\paragraph{Likelihood-based training signals.}
As a candidate RLVR-style training reward, the closest precedents are
works that use a likelihood-style signal on a held-out continuation $Y$
given a discrete auxiliary $Z$. VR-CLI~\cite{gurung2025longform} rewards
an auxiliary plan by per-token perplexity improvement on a gold
next-chapter continuation under a frozen reference model (a snapshot of
the policy at training start), in long-form story generation. In BOW~\cite{shen2025bow}, the model's entire reasoning trajectory is
given to the scorer; cf. our point (1) above. Gandhi et al.~\cite{gandhi2026dialogue}
optimize $\log p_\theta(Y \mid X, Z)$ on next-turn dialogue with the policy
serving as its own scorer; their finding that an LLM-as-judge reward
inflated judge scores while reducing true-utterance likelihood (worsening
when explicit reasoning is enabled) supports our emphasis on direct
likelihood of the held-out continuation rather than an explicit
natural-language judgment of whether $Z$ sounds useful. JEPO~\cite{tang2025jepo}
uses a variational (Jensen / ELBO) lower bound on log-prob to train on
math, including long-form proofs, treating the chain of thought as a
latent variable~\cite{hoffman2023cot,hu2024amortizing}.

Several other works explore variants of this signal as a training reward: NOVER~\cite{liu2025nover}
uses an EMA-style synchronized proxy of the policy as scorer; DRO~\cite{xu2025dro}
reweights per-token reference probabilities by their cross-rollout variance,
with rubric gates; VeriFree~\cite{zhou2025verifree} uses raw probability of
the reference answer given the reasoning trace; RLPR~\cite{yu2025rlpr} extends
the family to general non-verifiable domains. Quiet-STaR~\cite{zelikman2024quietstar}
and the pretraining family (RPT~\cite{dong2025rpt}, RLPT~\cite{li2025rlpt},
RLP~\cite{hatamizadeh2025rlp}) tie reasoning to future-token or
future-segment prediction at training time, with RLP closest
mathematically to our likelihood-ratio signal. Kwiatkowski et al.~\cite{kwiatkowski2026likelihood}
systematically compare log-prob, raw probability, and length-normalized
variants of reference-likelihood rewards, identifying log-probability of
the reference as most robust; our \cliptwo{} addresses a complementary failure
mode (catastrophic per-token losses dominating an otherwise reasonable
continuation).

\paragraph{Verifiers, scorer scale, and hackability.}
Process reward models and small-verifier work motivate treating the scoring
model as an object to be studied rather than as an oracle. Ref.~\cite{lightman2023verifyStep}
popularized step-level process supervision for mathematical reasoning;
Generative Verifiers frame reward modeling itself as next-token prediction
\cite{zhang2024generativeVerifiers}; prover-verifier games and verifier-scaling
studies show that verifier strength, generator strength, and task difficulty
interact in nontrivial ways
\cite{proverVerifier2024,variation2025}.

Reward-hackability concerns also shape our methodology: one-token judge
attacks, Reward Under Attack, verifier-gaming results, and reward-model
overoptimization all show that learned or model-based rewards can be
exploited
\cite{zhao2025onetoken,tiwari2026rewardattack,gamingVerifiers2025,
gao2023overoptimization,rafailov2024directOveroptimization,
inferenceRewardHacking2025}. These failures are not identical to our setting:
the forecast \(Z\) is not directly judged as a good answer. It is inserted into
a continuation prompt, and the scorer is evaluated on the true \(Y\). But the
analogous danger remains: a predictor might use \(Z\) for context stuffing,
format priming, scaffold exploitation, or other non-forecast strategies that
raise likelihood without carrying genuine predictive content. Our same-budget,
triple-budget, and context-only SFT controls are static stress tests for this
specific family of shortcuts.

\section{Limitations and discussion}

Several caveats apply to the interpretation of these results.

Our work here is a static benchmark and reward audit, not an RLVR training result. A predictor optimized directly against the reward could discover
strategies not represented by our current controls. The \cliptwo{} metric is a
defensible softened likelihood score, not a uniquely canonical objective. The
choice of scoring model matters, especially for longer continuations.
Cross-vendor predictor comparisons should also be interpreted carefully: the
placement of Opus 4.7 (medium reasoning) between GPT-5.5 (no reasoning) and
GPT-5.5 (low reasoning) is suggestive,
but it is not a calibrated capability ordering across providers or prompting
interfaces. Due to resource constraints, we did not run an Opus reasoning-effort
sweep analogous to the GPT-5.5 and GPT-5.4 nano sweeps. Finally,
equation-suffix prediction is a narrow technical forecasting task. It is
meaningful for scientific \TeX{} and mathematical reasoning, but the present
domain is equation-rich physics and mathematics rather than all technical
writing.

These limitations also point to the value of the benchmark. Because the task is
automatically generated and renewable, future work can add new recent papers, new
scoring-model families, and adversarially optimized controls. Future versions
could also implement heuristics for choosing which equation-continuation tasks
to score, especially if the goal is to sharpen adjacent reasoning-effort
comparisons. For example, as one heuristic, one could restrict to the 20\% of tasks on
which GPT-5.5 (no reasoning) has the lowest \cliptwo{} lift. We made this
observation in post-hoc exploration after collecting half the data, and it
holds qualitatively for the whole dataset: on the full benchmark, the
GPT-5.5 high-minus-low gap rises from \(+0.010\pm0.003\) to
\(+0.018\pm0.008\) (with Qwen scores), and from \(+0.011\pm0.003\) to
\(+0.018\pm0.008\) (with Kimi scores).

\section{Conclusion}

Likelihood lift provides a simple way to ask whether an auxiliary string helps a
fixed language-model scorer assign higher likelihood to a real hidden technical
continuation. In equation-suffix
prediction, forecast strings from strong models beat same-budget context
controls, rank model quality and reasoning effort, and survive a severe
context-only SFT stress test. This makes equation-suffix prediction a promising automatically generated
benchmark, which is moreover amenable to probes of shortcut vulnerabilities. Natural next steps are extensions to other technical domains, stronger
adversarial searches over non-forecast uses of the auxiliary string \(Z\), or
a small RL pilot.

\section*{Code and data availability}

Code, data artifacts, and reproduction scripts are available at
\url{https://github.com/danranard/arxiv-predictions}.

\section*{Acknowledgments}

The author designed the experiments and conducted the research; much of the code and some of the manuscript prose were drafted by OpenAI's GPT-5.5 and Anthropic's Claude Opus 4.7 under iterated revisions by the author.

\appendix

\section{Toy examples of equivalent equation continuations}
\label{app:equation-order-probe}

This appendix gives small controlled examples illustrating how log-probability
scoring behaves on distinct but mathematically equivalent continuations, and
the effect of \cliptwo{}.

Recall that in our equation-suffix scaffold, the scorer first reads a
completed ``first equation'' containing the auxiliary string \(Z\), then a
second copy of the same equation prefix, and only then scores the true target
suffix \(Y\) token by token. The forecast string itself is not scored.

The true target suffix is \(Y = X + A + B\). The exact-forecast condition
gives the scorer
\begin{shaded}
\centerline{$\displaystyle
\begin{array}{l}
\text{\% First equation:}\\
Z = X + A + B\\[0.25em]
\text{\% Same equation:}\\
Z =
\end{array}$}
\end{shaded}
\noindent and then scores the target \(X+A+B\). The reordered-forecast condition
instead gives
\begin{shaded}
\centerline{$\displaystyle
\begin{array}{l}
\text{\% First equation:}\\
Z = X + B + A\\[0.25em]
\text{\% Same equation:}\\
Z =
\end{array}$}
\end{shaded}
\noindent and still scores the same true target \(X+A+B\). The reordered forecast does
not match the paper's local \(A/B\) order, but it has still exposed the
scorer to the symbols \(A\) and \(B\) in the relevant equation context.
(In a real equation, an order mismatch of this kind may be mathematically
harmless or even exactly equivalent; the point here is only that it differs
from the particular continuation being scored.)

The key question is what happens after scoring passes the local mismatch. In
the reordered case, once the true scored prefix has reached
\(``\,Z = X + A +\,''\), the next token is \(B\). If the scorer still
assigns high probability to \(B\), then the reordered forecast differed
locally from the target but remained useful after recovery. The \cliptwo{} score is designed to reflect exactly this behavior: pay a bounded cost for the local
mismatch, but still count later tokens that became easy because the forecast
placed the scoring model in the right symbolic neighborhood.

We compare five conditions. We list them in the order of their empirical
\cliptwo{} scores; the ordering was the same under both scorers.
\begin{enumerate}
  \item exact forecast, where the first equation contains \(X+A+B\);
  \item reordered forecast, where the first equation contains \(X+B+A\)
        rather than the target order \(X+A+B\);
  \item wrong-symbol forecast, where the first equation contains \(X+Y\);
  \item context without explicit forecast, in which the scorer sees only the
        sentence ``Here are some equations involving some sums.'' before the
        equation prefix;
  \item empty scaffold, where no forecast string is inserted.
\end{enumerate}
The scored targets omit the closing display delimiter. Table~\ref{tab:qwen-addition-order-probe} shows the addition cases for the Qwen3-8B scorer; Kimi K2.6 gave the same qualitative ordering across both target cases. We also ran
the analogous juxtaposed-product cases \(X+AB\) versus \(X+BA\); they show
the same qualitative pattern, so we omit the table.

\begin{table}[p]
\centering
\scriptsize
\begin{tabular}{llrrr}
\hline
Case & Condition & Raw & \cliptwo{} & vs. empty \\
\hline
\(X+A+B\) & context (no forecast) & -1.949 & -1.103 & +0.440 \\
\(X+A+B\) & exact forecast & -0.008 & -0.008 & +1.534 \\
\(X+A+B\) & reordered forecast & -0.713 & -0.407 & +1.135 \\
\(X+A+B\) & wrong-symbol forecast & -3.238 & -0.900 & +0.642 \\
\(X+A+B\) & empty scaffold & -3.305 & -1.543 & +0.000 \\
\(X+B+A\) & context (no forecast) & -3.154 & -1.529 & +0.393 \\
\(X+B+A\) & exact forecast & -0.011 & -0.011 & +1.911 \\
\(X+B+A\) & reordered forecast & -0.869 & -0.438 & +1.484 \\
\(X+B+A\) & wrong-symbol forecast & -4.129 & -1.200 & +0.721 \\
\(X+B+A\) & empty scaffold & -4.241 & -1.922 & +0.000 \\
\hline
\end{tabular}
\caption{Toy equivalent-continuation examples, Qwen3-8B scorer. ``Raw'' is the mean per-token log-likelihood (the unsoftened limit of \cliptwo{}); \cliptwo{} is as defined in Section~\ref{sec:likelihood}. The ``vs.\ empty'' column gives \cliptwo{} differences from the empty-scaffold reference.}
\label{tab:qwen-addition-order-probe}
\end{table}

With the exact forecast in the scaffold, both scorers predict the later tokens of the target nearly deterministically. The reordered
forecast is punished at the local mismatch with the target, but it still beats
both the empty scaffold and the context-only condition by a large \cliptwo{} margin. This
supports the intended interpretation of \cliptwo{} as a more charitable score
for useful but nonliteral forecasts: it does not treat an order mismatch as an
exact continuation, but it also does not let that local mismatch erase later
predictive usefulness.

\begin{table}[p]
\centering
\scriptsize
\begin{tabular}{llrrrr}
\hline
Scorer & Case & Exact & Reordered & Wrong symbol & Empty \\
\hline
Qwen3-8B & \(X+A+B\) & 0.998 & 0.993 & 0.607 & 0.644 \\
Qwen3-8B & \(X+B+A\) & 0.996 & 0.857 & 0.074 & 0.033 \\
Kimi K2.6 & \(X+A+B\) & 0.987 & 0.987 & 0.507 & 0.705 \\
Kimi K2.6 & \(X+B+A\) & 0.989 & 0.970 & 0.045 & 0.089 \\
\hline
\end{tabular}
\caption{Probabilities the scorer assigns to the true next token at the \emph{recovery point} --- the position where the reordered forecast diverges from the target. For target \(X+A+B\) (rows 1, 3), this is \(P(B)\) after the partial target reaches \(X+A+\); for target \(X+B+A\) (rows 2, 4), \(P(A)\) after \(X+B+\). The columns vary which forecast was placed in the scaffold.}
\label{tab:addition-recovery-probe}
\end{table}

Table~\ref{tab:addition-recovery-probe} makes the same mechanism visible at the next-token level: once the true scored prefix has already passed the mismatch, the reordered forecast typically leaves the true next symbol highly probable.

\section{Sensitivity to the choice of softening}
\label{app:metric-sensitivity}

The headline metric \cliptwo{} floors each per-token log-likelihood at
\(-2\), but this is one specific choice among many bounded
transformations of the per-token score. To check that the qualitative
findings do not depend on this specific choice, we re-evaluate forecast
lift over the same-budget recent-context control under several
alternative softened scoring rules. Let
\(\lambda_t=\log p_J(y_t\mid\cdots)\) be the per-token log-likelihood; the
corresponding token loss is \(-\lambda_t\). All scores below are averaged over
target tokens, with higher values better.
We consider:
\begin{itemize}
  \item raw LL: $\lambda$ (no softening; the score equals the mean per-token log-likelihood).
  \item $\mathrm{clipLL}_k$: $\max(\lambda,-k)$, for
        $k \in \{2, 3, 5\}$. The headline \cliptwo{} is the case $k=2$.
  \item square-root loss score: $-\sqrt{-\lambda}$.
  \item log-one-plus loss score: $-\log(1-\lambda)$.
\end{itemize}
Each rule bounds the per-token contribution differently. Clipping floors
the per-token log-likelihood, equivalently capping the loss contribution; the
square-root and log-one-plus rules soften large losses smoothly without a
hard floor.

\begin{table}[h]
\centering
\scriptsize
\setlength{\tabcolsep}{3pt}
\begin{tabular}{lrrrrrrrrr}
\hline
Metric & \multicolumn{4}{c}{GPT-5.5} & \multicolumn{2}{c}{Opus 4.7} & \multicolumn{3}{c}{GPT-5.4 nano} \\
 & high & med & low & none & med & low & high & med & low \\
\hline
raw LL                    & $+0.134$ & $+0.132$ & $+0.117$ & $+0.073$ & $+0.106$ & $+0.091$ & $-0.048$ & $-0.061$ & $-0.089$ \\
\cliptwo{}                 & $+0.201$ & $+0.197$ & $+0.191$ & $+0.166$ & $+0.184$ & $+0.178$ & $+0.108$ & $+0.100$ & $+0.080$ \\
$\mathrm{clipLL}_3$       & $+0.227$ & $+0.223$ & $+0.215$ & $+0.183$ & $+0.207$ & $+0.198$ & $+0.106$ & $+0.098$ & $+0.072$ \\
$\mathrm{clipLL}_5$       & $+0.236$ & $+0.231$ & $+0.222$ & $+0.181$ & $+0.211$ & $+0.201$ & $+0.079$ & $+0.071$ & $+0.040$ \\
sqrt-loss score            & $+0.196$ & $+0.193$ & $+0.186$ & $+0.158$ & $+0.178$ & $+0.170$ & $+0.093$ & $+0.086$ & $+0.065$ \\
log-one-plus loss score    & $+0.129$ & $+0.126$ & $+0.121$ & $+0.102$ & $+0.116$ & $+0.111$ & $+0.054$ & $+0.049$ & $+0.034$ \\
\hline
\end{tabular}
\caption{Forecast lift over the same-budget recent-context control under
alternative softened scoring rules; Qwen3-8B scorer, 1363 equation cuts.
The qualitative ordering --- the GPT-5.5 reasoning-effort gradient is monotone,
GPT-5.5 with reasoning above Opus 4.7 above GPT-5.4 nano, within-Opus
and within-nano gradients also monotone --- holds under every bounded
variant. Under raw LL, GPT-5.4 nano lift becomes \emph{negative} on
this scorer: forecasts make the true continuation less likely than the
same-budget context control, illustrating the catastrophic-token
mechanism described in Section~\ref{sec:likelihood}.}
\label{tab:metric-sensitivity-qwen}
\end{table}

\begin{table}[h]
\centering
\scriptsize
\setlength{\tabcolsep}{3pt}
\begin{tabular}{lrrrrrrrrr}
\hline
Metric & \multicolumn{4}{c}{GPT-5.5} & \multicolumn{2}{c}{Opus 4.7} & \multicolumn{3}{c}{GPT-5.4 nano} \\
 & high & med & low & none & med & low & high & med & low \\
\hline
raw LL                    & $+0.129$ & $+0.121$ & $+0.107$ & $+0.069$ & $+0.103$ & $+0.090$ & $-0.020$ & $-0.031$ & $-0.058$ \\
\cliptwo{}                 & $+0.170$ & $+0.165$ & $+0.158$ & $+0.132$ & $+0.153$ & $+0.146$ & $+0.076$ & $+0.069$ & $+0.049$ \\
$\mathrm{clipLL}_3$       & $+0.179$ & $+0.173$ & $+0.166$ & $+0.133$ & $+0.160$ & $+0.151$ & $+0.062$ & $+0.054$ & $+0.030$ \\
$\mathrm{clipLL}_5$       & $+0.159$ & $+0.151$ & $+0.141$ & $+0.104$ & $+0.135$ & $+0.124$ & $+0.018$ & $+0.008$ & $-0.017$ \\
sqrt-loss score            & $+0.159$ & $+0.154$ & $+0.146$ & $+0.120$ & $+0.141$ & $+0.133$ & $+0.065$ & $+0.058$ & $+0.038$ \\
log-one-plus loss score    & $+0.107$ & $+0.103$ & $+0.098$ & $+0.079$ & $+0.094$ & $+0.089$ & $+0.038$ & $+0.033$ & $+0.018$ \\
\hline
\end{tabular}
\caption{Same as Table~\ref{tab:metric-sensitivity-qwen} but with the
Kimi K2.6 scorer.}
\label{tab:metric-sensitivity-kimi}
\end{table}

Across the substantially softened bounded variants, the qualitative findings are stable.
Under \cliptwo, $\mathrm{clipLL}_3$, the square-root loss score, and the
log-one-plus loss score, GPT-5.5
settings are monotone in reasoning effort under both scorers; GPT-5.5
with any reasoning sits above Opus 4.7, which sits above GPT-5.4 nano;
and the within-Opus and within-nano gradients are also monotone. The
specific numerical values shift but the orderings do not, so the choice
of \cliptwo{} as the headline metric is not load-bearing for these
conclusions.

Raw LL is an outlier, in exactly the way
Section~\ref{sec:likelihood} predicts. For GPT-5.4 nano under raw LL,
lift becomes negative on both scorers: forecasts make the true
continuation \emph{less} likely than the same-budget context control.
As the score is made less soft, the weakest nano settings begin to show
the same catastrophic-token behavior; this appears first for Kimi
$\mathrm{clipLL}_5$.
This is consistent with weaker forecasts occasionally producing a very
wrong token after a stretch where the scorer's predictive distribution
has sharpened, with that single sharpened-token mismatch dominating the
per-token average. The substantially softened variants restore positive lift,
confirming that the issue is concentrated in a small number of
catastrophic per-token losses rather than spread across the target.

\section{Two expanded equation-suffix examples}
\label{app:expanded-prompt-example}

This appendix shows two examples from the equation-suffix benchmark with the
main moving parts exposed: the predictor prompt, the true suffix, model
forecasts, the scaffolded scoring prompt, and the resulting scores. The first is
a clean win over the same-budget context control. The second is useful relative
to the empty scaffold, but does not beat the same-budget context control. The
actual predictor prompt contains roughly 10k characters of preceding paper
context. The excerpts shown below omit some parts, marked [omitted for brevity].

\subsection*{Example C.1: Clean useful forecast}

\boxheading{Source}
This example is from arXiv:2604.19885, \textit{On non-relativistic integrable
models and 4d SCFTs}. The hidden target is a suffix of an aligned displayed
equation.

\boxheading{Predictor prompt}
\begingroup
\small
\begin{shaded}
\begin{verbatim}
You are given recent context from a technical paper and the beginning of a
LaTeX display equation.
Continue the equation from exactly where it stops, in about 130 characters or fewer.
Write only the continuation. Do not write explanatory prose. Do not write \end{align}.

Recent paper context:
[omitted for brevity]

Let us first take the following ansatz for the eigenfunctions,
\begin{equation}
		[omitted for brevity]
\end{equation}
Here $J_\lambda(z)$ denotes the $A_1$ Jack polynomials, the well-known
eigenfunctions of the Lam\'{e} equation in the limit $q\rightarrow 0$.
The first few Jack polynomials take the following form.

Equation prefix:
\begin{align}
J^{(\alpha)}_0(z)&=1~,\nonumber\\
		J^{(\alpha)}_1(z)&=z+\frac{1}{z}~,\\
		J^{(\alpha)}_2(z)&=z^2+\frac{1}{z^2}+\frac{2 \alpha }{\alpha +
\end{verbatim}
\end{shaded}
\endgroup

\boxheading{True suffix}
\begingroup
\small
\begin{shaded}
\begin{verbatim}
1}~,\nonumber\\
		J^{(\alpha)}_3(z)&=z^3+\frac{1}{z^3}
		+\frac{3 \alpha }{\alpha +2}\left(z+\frac{1}{z}\right)\,,
		\cdots ~.\nonumber
\end{verbatim}
\end{shaded}
\endgroup
The actual scored target is this suffix followed by \verb|\end{align}|.

\boxheading{Returned forecasts}
GPT-5.5 (medium reasoning):
\begingroup
\small
\begin{shaded}
\begin{verbatim}
1}~,\nonumber\\
		J^{(\alpha)}_3(z)&=z^3+\frac{1}{z^3}
		+\frac{3 \alpha }{\alpha +2}\left(z+\frac{1}{z}\right)~,\nonumber\\
\end{verbatim}
\end{shaded}
\endgroup

GPT-5.4 nano (low reasoning):
\begingroup
\small
\begin{shaded}
\begin{verbatim}
1}~,\qquad J^{(\alpha)}_2(z)=z^2+\frac{1}{z^2}+\frac{2\alpha}{\alpha+1}\,,\nonumber\\
\end{verbatim}
\end{shaded}
\endgroup

\boxheading{Scoring prompt for GPT-5.5 (medium reasoning)}
The scoring prompt below ends at the second equation prefix. The target appended
and scored after this prompt is the true suffix plus \verb|\end{align}|.
\begingroup
\small
\begin{shaded}
\begin{verbatim}
% First equation:
\begin{align}
J^{(\alpha)}_0(z)&=1~,\nonumber\\
		J^{(\alpha)}_1(z)&=z+\frac{1}{z}~,\\
		J^{(\alpha)}_2(z)&=z^2+\frac{1}{z^2}+\frac{2 \alpha }{\alpha +1}~,\nonumber\\
		J^{(\alpha)}_3(z)&=z^3+\frac{1}{z^3}
		+\frac{3 \alpha }{\alpha +2}\left(z+\frac{1}{z}\right)~,\nonumber\\
\end{align}

% Same equation:
\begin{align}
J^{(\alpha)}_0(z)&=1~,\nonumber\\
		J^{(\alpha)}_1(z)&=z+\frac{1}{z}~,\\
		J^{(\alpha)}_2(z)&=z^2+\frac{1}{z^2}+\frac{2 \alpha }{\alpha +
\end{verbatim}
\end{shaded}
\endgroup

\boxheading{Scores}
\begin{center}
\begin{tabular}{lrrr}
\hline
Condition & \cliptwo{} score & vs. empty scaffold & vs. \(C_B\) \\
\hline
Empty scaffold & -0.402 & -- & -0.086 \\
\(C_B\) & -0.316 & +0.086 & -- \\
GPT-5.5 (medium reasoning) \(Z\) & -0.231 & +0.171 & +0.085 \\
GPT-5.4 nano (low reasoning) \(Z\) & -0.374 & +0.028 & -0.058 \\
\hline
\end{tabular}
\end{center}

This example is visually clean. GPT-5.5 (medium reasoning) predicts the next Jack-polynomial
structure with small formatting differences near the end. GPT-5.4 nano (low reasoning)
starts by closing the immediate denominator correctly, but then repeats the
\(J_2\) pattern instead of moving to \(J_3\).

\subsection*{Example C.2: Useful, but not above \(C_B\)}

This second example illustrates why the empty-scaffold and \(C_B\) comparisons
answer different questions. GPT-5.5 (medium reasoning) helps the scaffolded scorer relative
to no forecast, but the same-budget recent-context control is stronger on this
example.

\boxheading{Source}
This example is from arXiv:2604.24042. The hidden target is again a suffix of an
aligned displayed equation.

\boxheading{Predictor prompt}
\begingroup
\small
\begin{shaded}
\begin{verbatim}
You are given recent context from a technical paper and the beginning of a
LaTeX display equation.
Continue the equation from exactly where it stops, in about 130 characters or fewer.
Write only the continuation. Do not write explanatory prose. Do not write \end{align}.

Recent paper context:
[omitted for brevity]

Substituting the full nonlinear vector field with $\Delta=0$ and using
$y=a_3(t)x^3+O(x^5)$, the left-hand side becomes
\begin{align}
\dot{y}
&=
-\left(p(t)+\frac{\kappa}{2}\right)y-Kx(x^2+y^2) \notag\\
&=
-\left(p(t)+\frac{\kappa}{2}\right)a_3(t)x^3-Kx^3+O(x^5).
\end{align}
The right-hand side is

Equation prefix:
\begin{align}
\frac{\partial h}{\partial t}+\frac{\partial h}{\partial x}\dot{x}
&=
\dot{a}_3(t)x^3
+
3a_3(t)x^2
\left[
\left(p(t)-
\end{verbatim}
\end{shaded}
\endgroup

\boxheading{True suffix}
\begingroup
\small
\begin{shaded}
\begin{verbatim}
\frac{\kappa}{2}\right)x+Ky(x^2+y^2)
\right] \notag\\
&=
\dot{a}_3(t)x^3
+
3\left(p(t)-\frac{\kappa}{2}\right)a_3(t)x^3
+
O(x^5).
\end{verbatim}
\end{shaded}
\endgroup
The actual scored target is this suffix followed by \verb|\end{align}|.

\boxheading{Returned forecasts}
GPT-5.5 (medium reasoning):
\begingroup
\small
\begin{shaded}
\begin{verbatim}
\frac{\kappa}{2}\right)x+K y(x^2+y^2)
\right]+O(x^5) \notag\\
&=
\left[\dot{a}_3(t)+3a_3(t)\left(p(t)-\frac{\kappa}{2}\right)\right]x^3+O(x^5).
\end{verbatim}
\end{shaded}
\endgroup

GPT-5.4 nano (low reasoning):
\begingroup
\small
\begin{shaded}
\begin{verbatim}
\frac{\kappa}{2}\right)x-Kx(x^2+y^2)\right]
=3a_3(t)\left(p(t)-\frac{\kappa}{2}\right)x^3+O(x^5).
\end{verbatim}
\end{shaded}
\endgroup

\boxheading{Scoring prompt for GPT-5.5 (medium reasoning)}
As above, the scorer first sees the forecast inserted into a completed first
equation, then sees the same equation prefix again. The target appended and
scored after this prompt is the true suffix plus \verb|\end{align}|.
\begingroup
\small
\begin{shaded}
\begin{verbatim}
% First equation:
\begin{align}
\frac{\partial h}{\partial t}+\frac{\partial h}{\partial x}\dot{x}
&=
\dot{a}_3(t)x^3
+
3a_3(t)x^2
\left[
\left(p(t)-\frac{\kappa}{2}\right)x+K y(x^2+y^2)
\right]+O(x^5) \notag\\
&=
\left[\dot{a}_3(t)+3a_3(t)\left(p(t)-\frac{\kappa}{2}\right)\right]x^3+O(x^5).
\end{align}

% Same equation:
\begin{align}
\frac{\partial h}{\partial t}+\frac{\partial h}{\partial x}\dot{x}
&=
\dot{a}_3(t)x^3
+
3a_3(t)x^2
\left[
\left(p(t)-
\end{verbatim}
\end{shaded}
\endgroup

\boxheading{Scores}
\begin{center}
\begin{tabular}{lrrr}
\hline
Condition & \cliptwo{} score & vs. empty scaffold & vs. \(C_B\) \\
\hline
Empty scaffold & -0.566 & -- & -0.321 \\
\(C_B\) & -0.245 & +0.321 & -- \\
GPT-5.5 (medium reasoning) \(Z\) & -0.279 & +0.288 & -0.033 \\
GPT-5.4 nano (low reasoning) \(Z\) & -0.393 & +0.173 & -0.148 \\
\hline
\end{tabular}
\end{center}

Here the GPT-5.5 (medium reasoning) forecast captures the local continuation shape, while
GPT-5.4 nano (low reasoning) changes the sign and variable structure of the nonlinear term.
The example is nevertheless a loss against \(C_B\), which is precisely why the
main benchmark reports both weaker and stronger control contrasts.

\section{Examples of equation forecasts}
\label{app:equation-examples}
The examples below show equation-continuation tasks sampled from recent arXiv papers. In each example, the red marker indicates the point where the equation was hidden from the forecasting model. Each display aligns the paper continuation and the model forecast at a shared point before the hidden suffix, so differences after the cut are easier to compare.
For each example, the score line reports the \cliptwo{} likelihood gain from
conditioning on the forecast instead of a same-budget excerpt of immediately
preceding paper context. Positive values mean the forecast made the true hidden
suffix more likely under the scoring model.
The examples were sampled after the benchmark construction and scoring pipeline
were fixed, from cuts with available GPT-5.5 (high reasoning) forecasts and Qwen3-8B scores;
they were not selected for positive score.

\subsection*{Example 1}
\textit{Source:} \href{https://arxiv.org/abs/2604.24491}{arXiv:2604.24491}. \textit{Score gain over context baseline:} \(+0.171\).
\vspace{0.2em}
{\small
\[
\begin{aligned}
&\mathllap{\forall r\in\mathbb{N}^*\ ,\ s\in}{\color{red}\;|\!\mathrm{CUT}\!|\;}\frac{s_1+1}{2}+\mathbb{N}\ , \quad R_{r,s} = 0 \ . \qquad \text{\footnotesize paper}\\[0.45em]
&\mathllap{\forall r\in\mathbb{N}^*\ ,\ s\in}{\color{red}\;|\!\mathrm{CUT}\!|\;}\mathbb{N}^*\ ,\quad s\geq \frac{s_1+1}{2} \implies R_{r,s}=0 \ . \qquad \text{\footnotesize forecast}
\end{aligned}
\]
}
\vspace{0.4em}\hrule\vspace{0.8em}

\subsection*{Example 2}
\textit{Source:} \href{https://arxiv.org/abs/2604.18952}{arXiv:2604.18952}. \textit{Score gain over context baseline:} \(+0.338\).
\vspace{0.2em}
{\small
\[
\begin{aligned}
|\ell s| \gtrsim |s\partial_pA_{k-p,\ell-k+p}|
\ge |t\partial_p A_{k-p,p}| -{}
&{\color{red}\;|\!\mathrm{CUT}\!|\;}|\partial_p \Phi|
\ge \theta_0 |kt| - |\partial_p \Phi|
\ge \theta_0|kt|/2, \qquad \text{\footnotesize paper}\\[0.45em]
|\ell s| \gtrsim |s\partial_pA_{k-p,\ell-k+p}|
\ge |t\partial_p A_{k-p,p}| -{}
&{\color{red}\;|\!\mathrm{CUT}\!|\;}|\partial_p \Phi|
\ge \frac{\theta_0}{2}|kt|. \qquad \text{\footnotesize forecast}
\end{aligned}
\]
}
\vspace{0.4em}\hrule\vspace{0.8em}

\subsection*{Example 3}
\textit{Source:} \href{https://arxiv.org/abs/2604.19881}{arXiv:2604.19881}. \textit{Score gain over context baseline:} \(+0.136\).
\vspace{0.2em}
{\footnotesize
\[
\begin{aligned}
Y^{\mu}_{ab}Y_{\mu}^{ab}
&=\sum_{i=2,3}(\partial_a\partial_b \delta x_i-\tfrac{1}{2}\delta_{ab}\partial_c\partial^c\delta x_i)^2 \\[-0.1em]
\text{\footnotesize paper}\quad
&\tfrac{1}{2}\sum_{i={\color{red}\;|\!\mathrm{CUT}\!|\;}2,3}
\Big((\partial^2_{x_0}\delta x_i)^2+(\partial^2_{x_1}\delta x_i)^2
+2\,\partial^2_{x_1}\delta x_i\cdot \partial^2_{x_0}\delta x_i\Big)
=-\tfrac{1}{2}J(x_0,x_1),\\[0.45em]
\text{\footnotesize forecast}\quad
&\tfrac{1}{2}\sum_{i={\color{red}\;|\!\mathrm{CUT}\!|\;}2,3}
\Big((\partial^2_{x_0}\delta x_i-\partial^2_{x_1}\delta x_i)^2
+4(\partial_{x_0}\partial_{x_1}\delta x_i)^2\Big).
\end{aligned}
\]
}
\vspace{0.4em}\hrule\vspace{0.8em}

\subsection*{Example 4}
\textit{Source:} \href{https://arxiv.org/abs/2604.20635}{arXiv:2604.20635}. \textit{Score gain over context baseline:} \(+0.345\).
\vspace{0.2em}
{\small
\[
\begin{aligned}
&\mathllap{E=\frac{\partial\mathscr l}{\partial u}\cdot u-\mathscr l, \qquad p=}{\color{red}\;|\!\mathrm{CUT}\!|\;}\mathscr l-\rho\frac{\partial\mathscr l}{\partial\rho}. \qquad \text{\footnotesize paper}\\[0.45em]
&\mathllap{E=\frac{\partial\mathscr l}{\partial u}\cdot u-\mathscr l, \qquad p=}{\color{red}\;|\!\mathrm{CUT}\!|\;}\mathscr l-\rho\frac{\partial\mathscr l}{\partial \rho} \qquad \text{\footnotesize forecast}
\end{aligned}
\]
}
\vspace{0.4em}\hrule\vspace{0.8em}

\subsection*{Example 5}
\textit{Source:} \href{https://arxiv.org/abs/2604.19506}{arXiv:2604.19506}. \textit{Score gain over context baseline:} \(+0.090\).
\vspace{0.2em}
{\small
\[
\begin{aligned}
&\mathllap{N^{(2)}(z)=}\begin{cases} \mathcal{E}(z)N^{alg}(z), & z\in{\color{red}\;|\!\mathrm{CUT}\!|\;}\mathbb{C}\setminus\mathcal{D},\quad \mathcal{E}(z)N^{alg}(z)N^{\tilde{\kappa}}(z), z\in\mathcal{D}, \end{cases} \qquad \text{\footnotesize paper}\\[0.45em]
&\mathllap{N^{(2)}(z)=}\begin{cases} \mathcal{E}(z)N^{alg}(z), & z\in{\color{red}\;|\!\mathrm{CUT}\!|\;}\mathbb{C}\setminus\mathcal{D},\quad \mathcal{E}(z)N^{loc}(z), z\in\mathcal{D}. \end{cases} \qquad \text{\footnotesize forecast}
\end{aligned}
\]
}
\vspace{0.4em}\hrule\vspace{0.8em}

We show five examples here to keep the manuscript appendix compact.

\section{Numerical tables for main figures}
\label{app:figure-tables}

This appendix gives the numerical values plotted in Figures
\ref{fig:equation-benchmark-main}, \ref{fig:equation-static-controls}, and
\ref{fig:prose-window-lift}.
All entries use \cliptwo{} lift, reported per target token. Standard errors are
paper-clustered.

\begin{table}[p]
\centering
\scriptsize
\begin{tabular}{lrrrr}
\hline
Predictor setting & Qwen mean & Qwen SE & Kimi mean & Kimi SE \\
\hline
GPT-5.5 (high reasoning) & 0.201 & 0.006 & 0.170 & 0.006 \\
GPT-5.5 (medium reasoning) & 0.197 & 0.006 & 0.165 & 0.007 \\
GPT-5.5 (low reasoning) & 0.191 & 0.006 & 0.158 & 0.006 \\
GPT-5.5 (no reasoning) & 0.166 & 0.006 & 0.132 & 0.006 \\
Opus 4.7 (medium reasoning) & 0.184 & 0.006 & 0.153 & 0.006 \\
Opus 4.7 (low reasoning) & 0.178 & 0.006 & 0.146 & 0.006 \\
GPT-5.4 nano (high reasoning) & 0.108 & 0.006 & 0.076 & 0.006 \\
GPT-5.4 nano (medium reasoning) & 0.101 & 0.006 & 0.070 & 0.006 \\
GPT-5.4 nano (low reasoning) & 0.080 & 0.006 & 0.049 & 0.006 \\
\hline
\end{tabular}
\caption{Numerical values for the top panels of Figure
\ref{fig:equation-benchmark-main}: mean forecast lift over the same-budget
recent-context control \(\CB\). All rows use 1363 equation cuts from 138
papers.}
\label{tab:fig1-lift-values}
\end{table}

\begin{table}[p]
\centering
\scriptsize
\begin{tabular}{lrrrrrr}
\hline
Adjacent contrast & Qwen mean & Qwen SE & Qwen frac. + & Kimi mean & Kimi SE & Kimi frac. + \\
\hline
GPT-5.5 (high) -- (medium) & 0.004 & 0.003 & 0.389 & 0.005 & 0.002 & 0.519 \\
GPT-5.5 (medium) -- (low) & 0.006 & 0.003 & 0.411 & 0.006 & 0.003 & 0.537 \\
GPT-5.5 (low) -- (no reasoning) & 0.025 & 0.003 & 0.507 & 0.026 & 0.003 & 0.605 \\
Opus 4.7 (medium) -- (low) & 0.006 & 0.002 & 0.360 & 0.007 & 0.002 & 0.507 \\
GPT-5.4 nano (high) -- (medium) & 0.007 & 0.003 & 0.488 & 0.007 & 0.003 & 0.534 \\
GPT-5.4 nano (medium) -- (low) & 0.021 & 0.003 & 0.558 & 0.021 & 0.004 & 0.566 \\
\hline
\end{tabular}
\caption{Numerical values for the bottom panel of Figure
\ref{fig:equation-benchmark-main}: paired adjacent predictor-setting contrasts on
the same 1363 equation cuts. ``frac. +'' is the fraction of cuts on which the
paired difference is positive; parenthesized labels are reasoning-effort
settings.}
\label{tab:fig1-adjacent-values}
\end{table}

\begin{table}[p]
\centering
\scriptsize
\begin{tabular}{ll}
\hline
Predictor setting & Mean hidden or non-visible output tokens \\
\hline
GPT-5.5 (no reasoning) & 0 \\
GPT-5.5 (low reasoning) & 466.4 \\
GPT-5.5 (medium reasoning) & 1663.5 \\
GPT-5.5 (high reasoning) & 4789.9 \\
GPT-5.4 nano (low reasoning) & 376.0 \\
GPT-5.4 nano (medium reasoning) & 1518.5 \\
GPT-5.4 nano (high reasoning) & 2682.6 \\
Opus 4.7 (low reasoning) & \(\approx 101^\ast\) \\
Opus 4.7 (medium reasoning) & \(\approx 284^\ast\) \\
\hline
\end{tabular}
\caption{Generation-usage diagnostics for the equation-suffix predictor calls.
For OpenAI calls, hidden-reasoning tokens are API-reported reasoning tokens from
a representative generation run. The starred Opus 4.7 values are estimates:
Anthropic-reported total output tokens minus visible forecast tokens estimated by
the Anthropic \texttt{messages/count\_tokens} endpoint, with a small fixed
single-message overhead correction. The saved Anthropic usage records do not
expose the same hidden-vs-visible decomposition as OpenAI. Opus 4.7 may also
use tokenizer and accounting details not shared with earlier Claude models or
with OpenAI models. These numbers should therefore be read as scale diagnostics
for provider-defined effort settings, not as calibrated compute units across
providers.}
\label{tab:generation-usage-diagnostics}
\end{table}

\begin{table}[p]
\centering
\scriptsize
\begin{tabular}{lrrrr}
\hline
Condition & Mean vs. empty & SE & Cuts & Papers \\
\hline
Empty scaffold & 0.000 & 0.000 & 731 & 74 \\
Same-budget recent context \(\CB\) & 0.096 & 0.006 & 731 & 74 \\
Triple-budget recent context & 0.164 & 0.007 & 731 & 74 \\
GPT-5.5 forecast & 0.298 & 0.010 & 731 & 74 \\
True-suffix upper bound & 0.569 & 0.013 & 731 & 74 \\
Context-only SFT control & 0.200 & 0.017 & 220 & 25 \\
\hline
\end{tabular}
\caption{Numerical values for the control ladder in Figure
\ref{fig:equation-static-controls}. The first five conditions are evaluated on
the 731-cut subset used for this control audit. The context-only SFT condition
is evaluated on the held-out papers that overlap this figure's control subset
and were disjoint from its SFT training papers.}
\label{tab:fig2-control-ladder-values}
\end{table}

\begin{table}[p]
\centering
\scriptsize
\begin{tabular}{lrrrr}
\hline
Contrast & Mean & SE & Cuts & Papers \\
\hline
Forecast minus empty scaffold & 0.298 & 0.010 & 731 & 74 \\
Forecast minus same-budget context \(\CB\) & 0.202 & 0.008 & 731 & 74 \\
Forecast minus triple-budget context & 0.134 & 0.007 & 731 & 74 \\
Forecast minus context-only SFT & 0.082 & 0.013 & 220 & 25 \\
True-suffix upper bound minus forecast & 0.271 & 0.010 & 731 & 74 \\
\hline
\end{tabular}
\caption{Numerical values for the right panel of Figure
\ref{fig:equation-static-controls}: paired forecast advantages over the control
conditions. The SFT condition is evaluated on the held-out papers that overlap
this figure's control subset and were disjoint from its SFT training papers; the
other conditions use the 731-cut subset used for this control audit.}
\label{tab:fig2-forecast-control-values}
\end{table}

\begin{table}[p]
\centering
\scriptsize
\begin{tabular}{lrrrr}
\hline
Predictor setting & 50 tokens & 100 tokens & 200 tokens & 400 tokens \\
\hline
GPT-5.5 (no reasoning) & \(0.033\pm0.003\) & \(0.025\pm0.003\) & \(0.017\pm0.003\) & \(0.009\pm0.002\) \\
GPT-5.5 (low reasoning) & \(0.036\pm0.004\) & \(0.028\pm0.003\) & \(0.019\pm0.002\) & \(0.011\pm0.002\) \\
GPT-5.5 (medium reasoning) & \(0.033\pm0.004\) & \(0.027\pm0.003\) & \(0.020\pm0.003\) & \(0.013\pm0.002\) \\
GPT-5.5 (high reasoning) & \(0.035\pm0.004\) & \(0.028\pm0.003\) & \(0.019\pm0.003\) & \(0.011\pm0.002\) \\
GPT-5.4 nano (low reasoning) & \(0.003\pm0.003\) & \(0.003\pm0.002\) & \(0.000\pm0.002\) & \(-0.002\pm0.001\) \\
GPT-5.4 nano (medium reasoning) & \(0.008\pm0.004\) & \(0.006\pm0.003\) & \(0.002\pm0.002\) & \(-0.001\pm0.001\) \\
GPT-5.4 nano (high reasoning) & \(0.013\pm0.003\) & \(0.008\pm0.002\) & \(0.004\pm0.002\) & \(0.001\pm0.001\) \\
\hline
\end{tabular}
\caption{Numerical values for Figure \ref{fig:prose-window-lift}: forecast
lift over the same-budget recent-context control in the prose/\TeX{}
continuation experiment, shown as mean \(\pm\) paper-clustered SE for each scored
target-token window. GPT-5.5 rows use 654 cuts except GPT-5.5 (high reasoning), which uses
639; GPT-5.4 nano rows use 651 cuts. All rows use 40 papers.}
\label{tab:fig3-prose-window-values}
\end{table}

\clearpage
\printbibliography

\end{document}